\documentclass{article}

\usepackage{PRIMEarxiv}

\usepackage[utf8]{inputenc} 
\usepackage[T1]{fontenc}    
\usepackage{hyperref}       
\usepackage{url}            
\usepackage{booktabs}       
\usepackage{amsfonts}       
\usepackage{nicefrac}       
\usepackage{microtype}      
\usepackage{lipsum}
\usepackage{fancyhdr}       
\usepackage{graphicx}       
\graphicspath{{media/}}     

\usepackage{hyperref}
\usepackage{multirow}
\usepackage{amsmath} 
\usepackage{float}

\title{Graph RAG as Human Choice Model: Building a Data-Driven Mobility Agent with Preference Chain
}

\author{
  Kai Hu \\
  School of Architecture \\
  South China University of Technology \\
  Guangzhou, China\\
  \texttt{arhukai@mail.scut.edu.cn} \\
   \And
  Parfait Atchade-Adelomou \\
  MIT Media Lab \\
  Massachusetts Institute of Technology \\
  Cambridge, MA, USA\\
  \texttt{parfait@media.mit.edu} \\
  \And
  Carlo Adornetto \\
  Dept. of Mathematics and Computer Science \\
  University of Calabria \\
  Via Pietro Bucci, Rende,Italy\\
  \And
  Adrian Mora-Carrero \\
  Escuela Técnica Superior de Ingenieros Informáticos \\
  Universidad Politécnica de Madrid \\
  Campus Montegancedo, Madrid, Spain\\
  \And
  Luis Alonso-Pastor \\
  MIT Media Lab \\
  Massachusetts Institute of Technology \\
  Cambridge, MA, USA\\
  \And
  Ariel Noyman \\
  MIT Media Lab \\
  Massachusetts Institute of Technology \\
  Cambridge, MA, USA\\
  \And
  Yubo Liu \\
  School of Architecture \\
  South China University of Technology \\
  Guangzhou, China\\
  \And
  Kent Larson \\
  MIT Media Lab \\
  Massachusetts Institute of Technology \\
  Cambridge, MA, USA
}

\begin{document}
\maketitle

\begin{abstract}

Understanding human behavior in urban environments is a crucial field within city sciences. However, collecting accurate behavioral data, particularly in newly developed areas, poses significant challenges. Recent advances in generative agents, powered by Large Language Models (LLMs), have shown promise in simulating human behaviors without relying on extensive datasets. Nevertheless, these methods often struggle with generating consistent, context-sensitive, and realistic behavioral outputs. To address these limitations, this paper introduces the Preference Chain, a novel method that integrates Graph Retrieval-Augmented Generation (RAG) with LLMs to enhance context-aware simulation of human behavior in transportation systems. Experiments conducted on the Replica dataset demonstrate that the Preference Chain outperforms standard LLM in aligning with real-world transportation mode choices. The development of the Mobility Agent highlights potential applications of proposed method in urban mobility modeling for emerging cities, personalized travel behavior analysis, and dynamic traffic forecasting. Despite limitations such as slow inference and the risk of hallucination, the method offers a promising framework for simulating complex human behavior in data-scarce environments, where traditional data-driven models struggle due to limited data availability.

\end{abstract}


\section{Introduction}

Studying human activities in urban environments has emerged as a critical approach for informed decision-making in transportation systems, particularly as cities grapple with increasing complexity, resource constraints, and the need for adaptive infrastructure\cite{zhangDataDrivenIntelligentTransportation2011}. Traditional agent-based models (ABMs) rely on predefined rule sets to replicate human activities, yet their rigid frameworks often fail to capture the inherent uncertainty and variability in real-world mobility patterns\cite{matthewsAgentbasedLanduseModels2007}. While machine learning (ML) and deep learning (DL) methods have been increasingly employed to predict complex human behaviors\cite{fuchs2023modeling}, they frequently face limitations in adaptability to dynamic conditions or require extensive, high-quality datasets—resources that are often scarce in emerging urban contexts. 

Recent advances in LLMs have shown promise in simulating human-like actions. Researchers have explored the use of generative agents to replicate complex behaviors in digital environments without the need for additional data\cite{adornetto2025generative}. Despite their potential, LLMs often produce outputs that lack consistency or fail to provide contextually appropriate responses\cite{novikova2025consistency}. The absence of contextual guidance during the generative process can result in behaviors that do not accurately reflect real-world dynamics, thereby limiting their utility as urban planning and analysis tools.

This paper addresses these challenges by introducing Preference Chain, a combined approach leveraging Graph RAG and LLMs to enhance the human behavior modeling in transportation systems. In the proposed framweork, Graph RAG employs graph-based retrieval to incorporate structured knowledge, enhancing the contextual accuracy of the model. Meanwhile, the LLM contributes general knowledge of human behaviors, enabling the model to adapt effectively to diverse scenarios. By integrating the Preference Chain within an autonomous traffic simulation agent named the Mobility Agent, this paper demonstrate its capacity to replicate realistic urban mobility patterns with limited data. 

The experiments indicate that the proposed method improves the realism and reliability of LLM simulated behaviors across population groups, the proposed method offers city scientists and planners a valuable tool which supports behavior simulation tasks in data-scarce regions such as urban mobility modeling, personalized travel behavior analysis, and dynamic traffic forecasting.

\section{Related Work}
\label{sec:related_work}

\subsection{Agent-based Model}
An ABM is a computational method for modeling complex systems and processes. By aggregating the decisions of autonomous agents within a simulated environment, this approach provides both descriptive and predictive insights into real-world dynamics. ABMs construct systems from the bottom up, enabling individual agents' autonomous and social characteristics to interact in complex, nonlinear ways. ABMs have been widely used in ecological and environmental modeling\cite{mamy2024comparison}, transportation simulation\cite{coretti2023urban}, and urban and architectural designs\cite{alma9935455324406761}. Traditional ABMs capture agent behaviors and interactions' inherent uncertainty and variability through rule-based, probabilistic, and logical paradigms\cite{chenReviewApplicationsAgent2010}. However, defining such complex behaviors within a fixed set of rules can be challenging.

\subsection{Human Behavior Prediction}
With advancements in computer science, researchers have tried to analysis and predict human behaviors in diverse contexts using ML approaches. For example, Poongodi et al. used XGBoost to predict taxi trips\cite{poongodiNewYorkCity2022}, while Kamsiriochukwu et al. showed MLPs are effective for activity recognition \cite{app132011154}. Inigo and Eduardo Bilbao applied Random Forest to evaluate mobility interventions \cite{azcarateurrutia2024mode,pavon2025predicting}. These methods work well with relatively small datasets but require extensive feature engineering and may lack adaptability in changing conditions. DL techniques, on the other hand, such as LSTM \cite{tang2021trip}, Transformers\cite{kobayashi2023modeling}, and GCN\cite{zhaoTGCNTemporalGraph2020}, offer improved performance by capturing complex temporal and spatial patterns.  However, these models typically require large-scale behavior data collected from (often millions of) mobile devices or social media platforms, which is often costly and may raise privacy concerns.

\subsection{Generative Agent}
Recent research has shown that LLMs encode a wide range of human behaviors derived from their training data\cite{brown2020language}. By integrating LLMs with ABM, generative agents can produce reasonable behavioral patterns. For instance, Park \textit{et al.}  presents an architecture for generative agents that can communicate with each other and interact with the environment\cite{10.1145/3586183.3606763}. Kaiya \textit{et al.} introduces a highly autonomous generative agent designed to simulate complex social behaviors in virtual societies with low computational cost and real-time human interaction\cite{kaiya2023lyfeagentsgenerativeagents}. Atchade \textit{et al.} proposes a Humanized Agent-Based Models (h-ABM) framework that integrates LLMs into ABM to simulate human-like behaviors, emotions, and decision-making in complex systems, offering a modular approach for enhanced realism and applications in urban planning and beyond\cite{atchade-adelomouHumanizedAgentbasedModels}.

Despite these advancements, generative agents still face several challenges. One major issue is the problem of "hallucination" in LLM response\cite{mirzadeh2024gsmsymbolicunderstandinglimitationsmathematical}, where LLMs produce inaccurate or fabricated information, compromising the reliability of generative agents. Additionally, although LLMs are trained on general datasets that capture broad human behaviors, they lack the localized knowledge essential for specific contexts. In urban studies, understanding the local context is crucial, as generic knowledge may not suffice for accurately representing place-specific behaviors.

\subsection{Graph Retrieval Augmented Generation}
RAG is a framework that combines information retrieval with the generation capabilities of language models\cite{lewis2020retrieval}. It is used to enhance the quality of LLM responses by retrieving relevant data from external sources of knowledge. RAG is an efficient and effective method for providing LLMs with local contextual knowledge without fine-tuning. Graph RAG is an approach that builds upon RAG to address the limitations of RAG, where both textual and topological information are important. It emphasizes the importance of subgraph structures to enhance the retrieval and generation processes\cite{edgeLocalGlobalGraph2024}.

Based on prior research in human behavior modeling, this paper addresses the limitations of existing approaches in capturing realistic, context-sensitive simulations within limited data. Behavior modeling in traditional ABMs often rely on predefined rule sets, which makes it challenging to accommodate complex human behaviors. ML/DL models offer an alternative by predicting human actions. Yet, these methods typically demand large datasets for accuracy. While generative agents based on LLMs can emulate complex human behaviors without additional data, they often struggle with hallucination and context-specific awareness.

This paper proposes a combined approach integrating Graph RAG with LLMs. By incorporating structured, contextually relevant information through graph-based retrieval, this approach provides LLM agents with the nuanced understanding necessary to simulate realistic, localized human choices and the capability to adapt to diverse scenarios.

\section{Methodology} \label{sec:methodology}
Chain of Thought (CoT)\cite{wei2022chain} and Tree of Thought (ToT)\cite{yao2023tree} are two commonly used techniques for LLMs to generate more reasonable responses. CoT improves reasoning ability by decomposing complex problems into a series of intermediate steps, while ToT enhances the breadth and depth of reasoning by exploring multiple possible solution paths. However, both methods rely on explicit logic prompts---i.e., predefined logical structures that guide the model's reasoning process. However, human behavior simulation, in this case, transportation behavior, presents unique challenges due to its inherent complexity and variability under different conditions. Behavioral patterns can change significantly depending on the context, making it extremely difficult to guide an LLM's reasoning about behavior choices across different regions and individuals using explicit logic prompts.

\begin{figure*}[!ht]
    \centering
    \includegraphics[width=\linewidth]{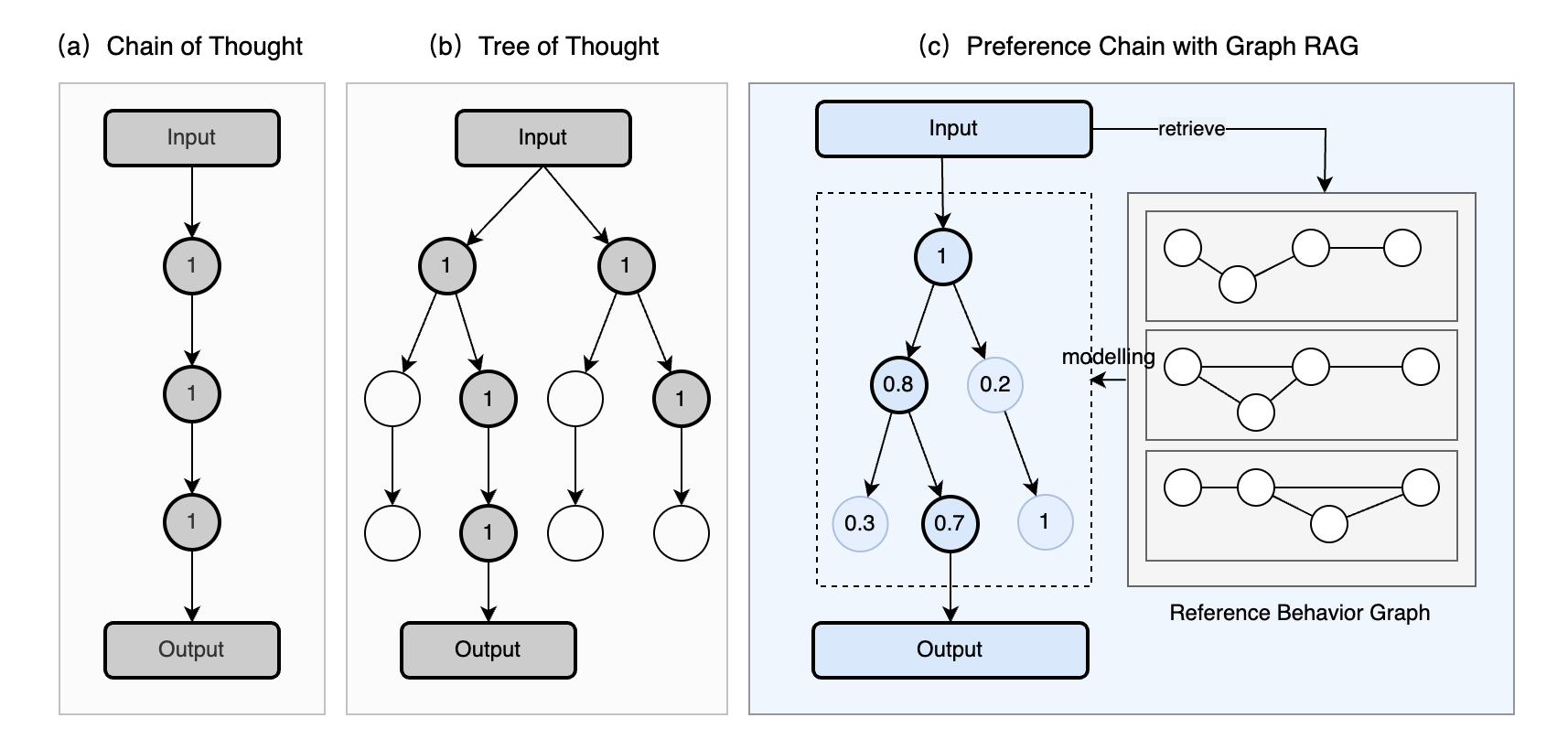}
    \caption{CoT,ToT vs Preference Chain}
    \label{fig:preferench_chain}
\end{figure*}

As shown in Figure \ref{fig:preferench_chain}, this paper proposes a method named Preference Chain, which leverages Graph RAG to construct individual behavior preference from a small amount of data. Unlike the fixed reasoning logic of CoT and ToT, the Preference Chain builds dynamic action sequences based on the dataset. The type of behavior and the probability of selecting different behaviors in a given situation can dynamically vary depending on the individual profiles and the relevant data. Through this approach, the Preference Chain can more flexibly and finely simulate individual behavior preferences, thereby providing effective guidance for LLMs in simulating complex behaviors.

There are four steps to construct a Preference Chain:
creating a behavioral graph, similarity search, behavioral probabilistic modeling, and LLM preferences remodeling:

\begin{figure}[!hb]
    \centering
    \includegraphics[width=0.8\linewidth]{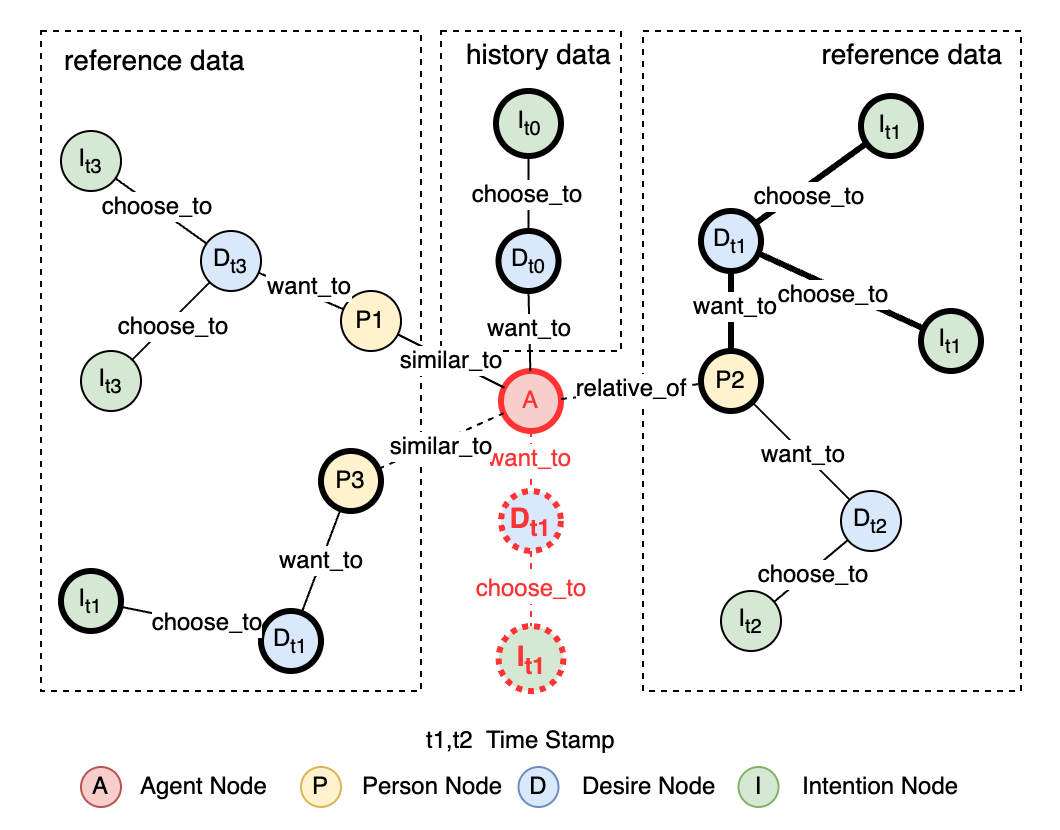}
    \caption{The schema of a behavior graph}
    \label{fig:behavior graph}
\end{figure}

\subsection{BDI Transportation Behavior Graph}
The first step involves the construction of a graph database following  the \textit{Belief-Desire-Intension} (BDI) architecture to model and analyze the relationships between human profiles and transportation behaviors, thereby enabling a comprehensive understanding of behavioral patterns and informing the simulation of LLM behaviors. The BDI architecture, rooted in the philosophy of action\cite{bratman1987intention}, represents how agents perform complex reasoning by encoding their knowledge, goals, and actions. This architecture has been widely adopted in simulation platforms such as GAMA\cite{caillou2017simple} and Netlogo\cite{sakellariou2008enhancing}.

As shown in Figure \ref{fig:behavior graph}, The behavior graph is represented as a weighted directed graph $G=(V,E,\omega)$, which includes agents, potential intentions, and related contextual information. Specifically:
\begin{itemize}
    \item $V$ is the set of nodes, consisting of Agent, Person (similar individuals), Desire (similar needs), and Intention (potential intentions).
    \item $E$ is the set of edges, with relationship types including $relative\_of$ (family/social relationships), $similar\_to$, $want\_to$, and $choose\_to$.
    \item $\omega$ denotes the weight $\omega(e)$ of each edge $ e \in E$. All weights are in the range $[0, 1]$, and the specific meaning of each weight is determined by the type of edge: $\omega_\text{rel}$ represents the closeness of family or social relationships; $\omega_\text{sim}$ represents the similarity of personal attributes;$\omega_\text{want}$ represents the similarity of desires;$\omega_\text{choice}$ represents the temporal proximity of choices.
\end{itemize}

\subsection{Similarity Search}
In order to provide more accurate and effective behavioral reference information for the LLM agent, this paper uses similarity search to identify the most similar people to the current simulated agent, as well as the choices they made when facing similar situation. The similarity search algorithm consists of two steps: vector similarity search and depth-first graph search.

\begin{itemize}
    \item Vector Similarity Search: This paper employs the
    \textit{mxbai-embed-large}\cite{emb2024mxbai} model to convert the profile texts of humans and agents into embedding vectors. The model has learned rich semantic information, enabling the generated embeddings to effectively reflect the semantic content. By computing the cosine similarity between vectors, it is possible to efficiently identify individuals whose profiles are most similar to those of the current simulated agent.
    \item Depth-first Graph Search: After identified the most similar person nodes. A depth-first graph search (depth=3) is applied for each person node to collect the related desire nodes and intention nodes for constructing the behavioral subgraph. This strategy aims to effectively uncover potential behavioral relationships that are relevant to the current agent, while also limiting the search scope to improve computational efficiency.
\end{itemize}

\subsection{Probabilistic Modeling}
After constructing a subgraph that includes the simulated agent, similar humans, and their past behaviors, a link prediction algorithm is used to model the probability of the agent choosing among various options. This paper employ a path analysis method based on behavioral graphs to model the probability of different choices under the current state of an agent. For an agent $a$ and a specific intention $i$, the process of calculating the selection probability is described as follows:

Let $\mathcal{P}_{a \to i}^{(K)}$ denote the set of all simple paths from the agent node $a$ to the intention node $i$. For any path $p = (e_1, e_2, \dots, e_k)$ in the set, where $k \leq K$ and $e_j$ represents the $j$-th edge in the path, the weight $W(p)$ of the path is defined as the product of the weights of its constituent edges:

\begin{equation}
W(p) = \prod_{j=1}^{k} \omega(e_j)
\end{equation}

For example, if a specific path $p'$ sequentially traverses edges representing relationship closeness ($\omega_{\text{rel}}$), profile similarity ($\omega_{\text{sim}}$), desire similarity ($\omega_{\text{want}}$), and temporal proximity ($\omega_{\text{choice}}$), then the weight of path $p'$ can be expressed as the product of these edge weights:

\begin{equation}
W(p') = \omega_{\text{rel}} \cdot \omega_{\text{sim}} \cdot \omega_{\text{want}} \cdot \omega_{\text{choice}}
\end{equation}

The raw preference score $S(a,i)$ of agent $a$ for intention option $i$ is defined as the sum of the weights of all valid paths $p \in \mathcal{P}_{a \to i}^{(K)}$:

\begin{equation}
S(a,i) = \sum_{p \in \mathcal{P}_{a \to i}^{(K)}} W(p)
\end{equation}

Finally, the raw preference scores for each intention option are normalized to compute the relative likelihood (selection probability) $P(i|a)$ of agent $a$ choosing intention $i$:

\begin{equation}
P(i|a) = \frac{S(a,i)}{\sum_{j \in \mathcal{I}} S(a,j)}
\end{equation}

where $\mathcal{I}$ denotes the set of all candidate intention options.

This method leverages the node information and edge weights in the behavioral subgraph, capturing the behavioral similarity and preference patterns between the agent and individuals with similar profiles and desires. Compared to other methods (such as ML or DL), this approach does not require a complex training process or a large dataset, making it particularly suitable for scenarios in dynamic simulations with limited data.

\subsection{LLM preferences remodeling}

Probabilistic modeling with link prediction can only provide information about what happened under past conditions, which reduces its adaptability to new situations. LLMs have demonstrated significant potential for simulating human behavior and responding to diverse conditions\cite{brown2020language,10.1145/3586183.3606763}. 

In this paper, the initial probabilities of the link prediction are treated as a prior distribution. The LLM is then leveraged to refine and calibrate these probabilities based on newly introduced environmental conditions. These conditions are described in natural language and may include information such as weather, time, or city, thereby coupling the model's
decision-making process with real-world contextual factors. By integrating the structured information from link prediction algorithms with the semantic
reasoning capabilities of LLMs, this framework not only predicts conventional behavioral patterns but also effectively handles complex scenarios that were not present in the original dataset, thereby achieving more realistic behavioral simulation.

\section{Experiment I: Transportation Mode Simulation}
\label{sec:exp_1}

To evaluate the performance of the proposed method in improving the accuracy of LLMs in behavior simulation, this paper conducts an experiment on simulating modes of transportation.

\subsection{Datasets}
This experiment utilizes the mobility data from Replica\cite{replica2025}. This data represents traffic trips for a typical Thursday during Spring 2024 in Cambridge, MA and San Francisco, CA. Reference data samples (range from 10 to 1000) and another 1000 validation samples are randomly selected from the original dataset. The Replica dataset contains key information such as traveler attributes (age, income, employment status, household size, number of vehicles owned, and education level),trip start time, trip purpose, trip duration, transportation mode. All data have been anonymized using ML techniques to protect the privacy of real individuals. To simplify the model, continuous variables in the Replica dataset, including age, income, trip start time, and trip duration, were categorized. The schema and more descriptive analysis of the dataset is provided in the Appendix \ref{appendix:data_schema}.


\subsection{Compared Models}

The primary objective of this paper is to improve the accuracy of LLM in simulation of human behavior with limited data. The experiment compares the performance of LLM approach and the proposed Preference Chain approach. Unless specified, the Qwen3:8b\cite{qwen3technicalreport} model without thinking mode is used for both LLM and Preference Chain simulation in this paper. Several popular ML algorithms, including Random Forest (RF) and Multilayer Perceptron (MLP), implemented using the scikit-learn library\cite{pedregosa2011scikit}, as well as XGBoost (XGB)\cite{chen2016xgboost}, are employed to compare with the proposed method. Since DL methods typically require large-scale datasets, which are beyond the scope of this study, this experiment did not include DL methods. Detailed information on the compared models can be found in the Appendix \ref{appendix:compared_models}.

\subsection{Metrics}
Commonly used metrics in human behavior modeling include distance metrics, classification metrics, error metrics, similarity metrics, and divergence metrics\cite{luca2021survey}. This study employs Kullback-Leibler Divergence (KLD) and Mean Average Error (MAE) to assess the distribution similarity and prediction accuracy of different methods.

Kullback-Leibler divergence (KLD): The KLD measures how different a probability distribution is from a reference probability distribution. The formula for KLD is:
\begin{equation}
\text{KLD} = \sum_{i} \sum_{j} P(i, j) \log \left( \frac{P(i, j)}{Q(i, j)} \right)
\end{equation}
where:$P(i,j)$ is the true joint probability distributions between population groups$ i $ and choices$ j $, $Q(i,j)$is the predicted probability distribution.

Mean Average Error (MAE): MAE calculates the average of the absolute differences between the predicted probability $ Q(i,j) $ and the actual probability $ P(i,j) $ for different population groups and choices. A lower MAE indicates a better fit between the model’s predictions and the actual values. The formula for MAE is:
\begin{equation}
\text{MAE} = \frac{1}{N} \sum_{i} \sum_{j} \left| P(i,j) - Q(i,j) \right|
\end{equation}
where: $ P(i,j) $ is the true probability of population group $ i $ choosing option $ j $; $ Q(i,j) $ is the predicted probability of population group $ i $ choosing option $ j $.

\subsection{Evaluation}

\textbf{1. Does the proposed method perform better than the LLM approach with limited data?}

\begin{figure}[!h]
    \centering
    \includegraphics[width=\linewidth]{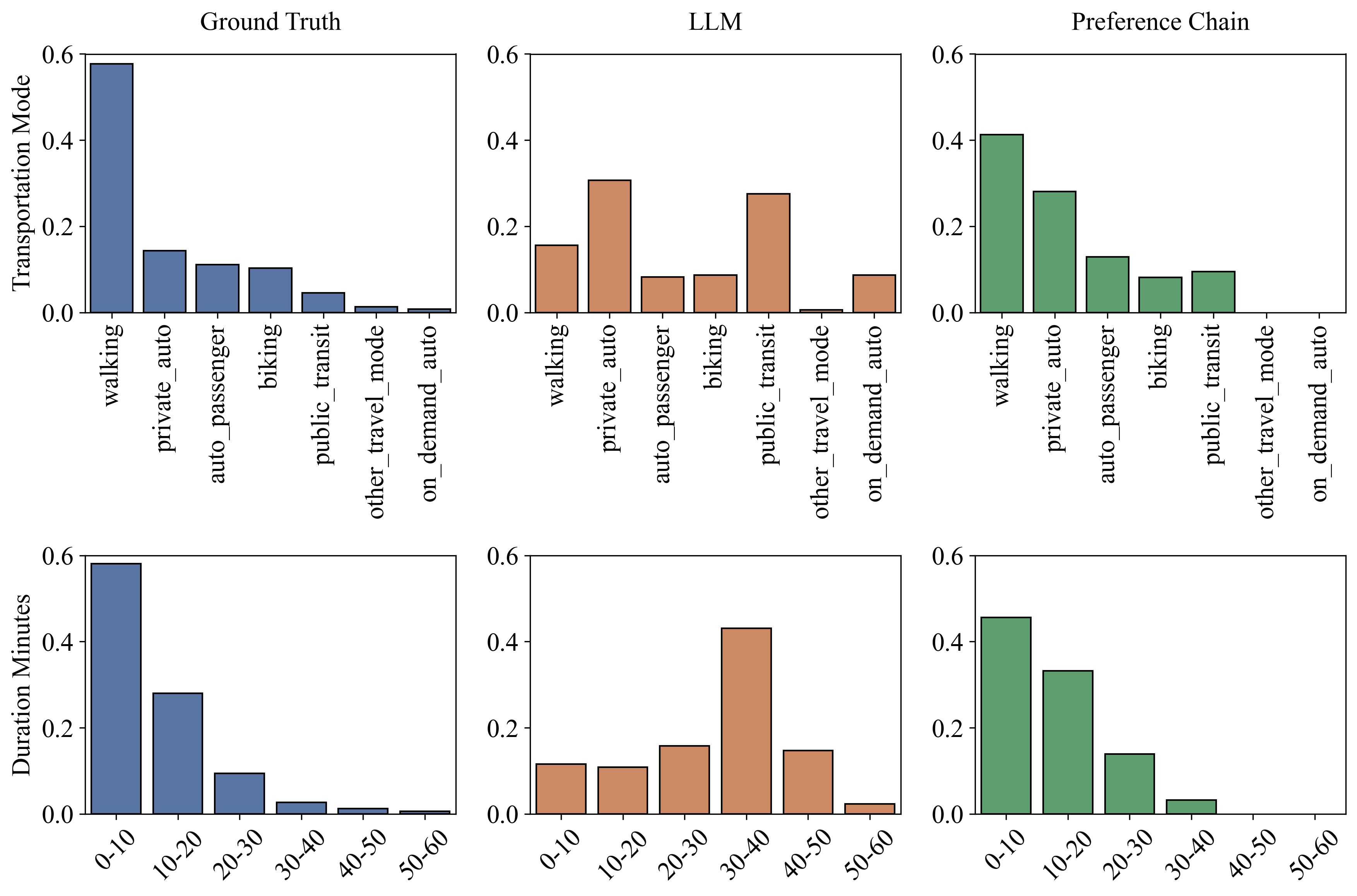}
    \caption{The predicted distribution of the Preference Chain is closer to the ground truth than that of the LLM method}
    \label{fig:preference_chain_vs_ground_truth}
\end{figure}

The experiment simulates mode choices and trip duration choices of 1,000 individuals with different profiles to evaluate the performance of proposed method. 50 samples not included in the validation dataset are randomly selected for Preference Chain as the reference data. As shown in Figure \ref{fig:preference_chain_vs_ground_truth}, The proposed method indeed makes the prediction distribution more consistent with the real situation. However, it is worth noting that the Preference Chain also caused the model to ignore less frequently chosen options, such as "on-demand car services" and "other travel modes", as well as travel times exceeding 40 minutes.

\begin{figure}[!h]
    \centering
    \includegraphics[width=\linewidth]{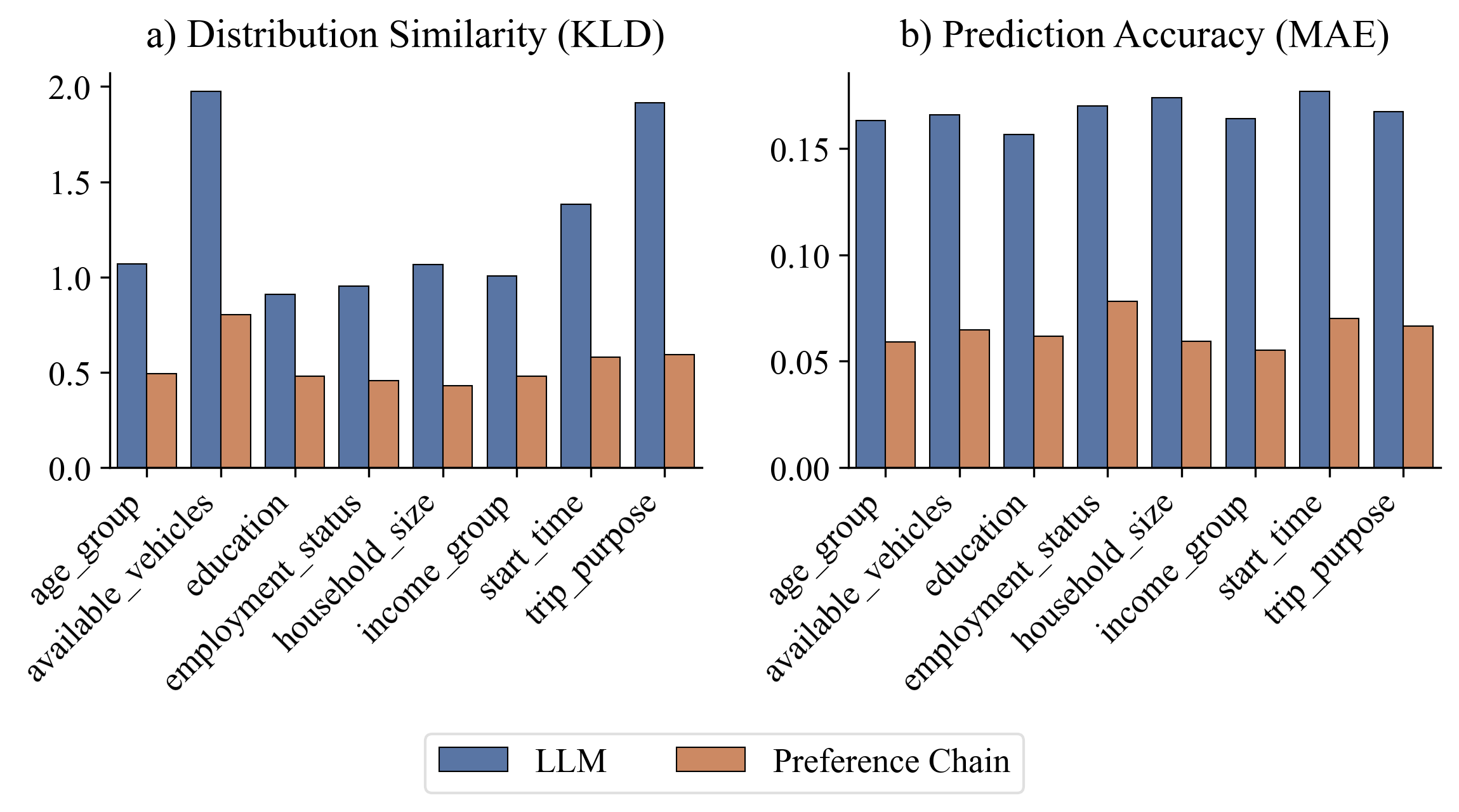}
    \caption{Preference Chain outperforms LLM approach across all demographic groups}
    \label{fig:preference_chain_vs_llm}
\end{figure}

Further comparison of the distribution similarity (KLD) and prediction accuracy (MAE) between the simulated data and the real data across different dimensions reveals that the Preference Chain approach significantly improves the accuracy of LLMs across all dimensions (See Figure \ref{fig:preference_chain_vs_llm}). This indicates that the Preference Chain can enable LLMs to better simulate the travel behavior of individuals with different profile.


\textbf{2. How much reference data are needed to construct the Preference Chain?}

\begin{figure}[!ht]
    \centering
    \includegraphics[width=\linewidth]{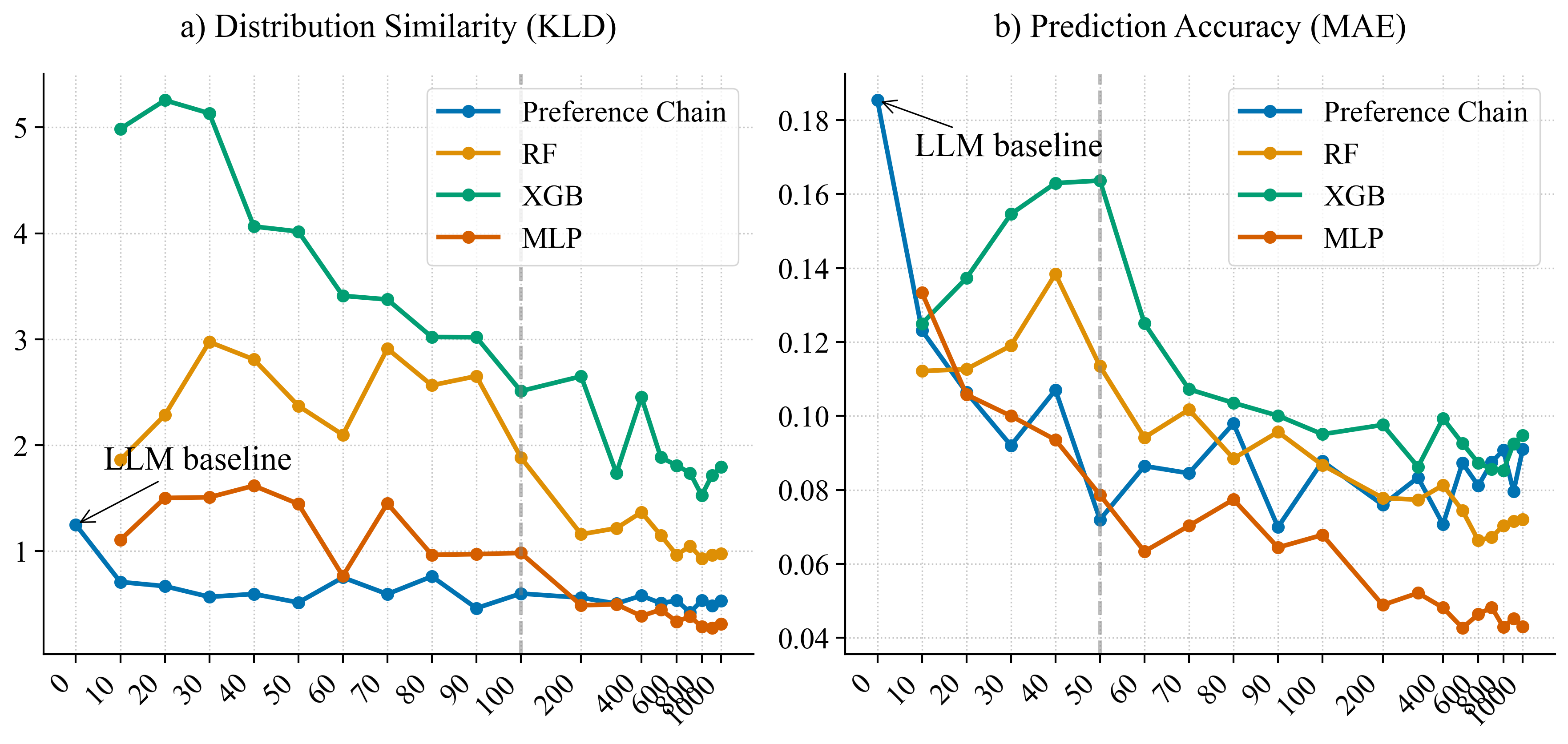}
    \caption{The Preference Chain consistently outperforms the LLM approach, and outperform other ML methods within approximately 50 reference/train samples.}
    \label{fig:preference_chain_vs_ml}
\end{figure}

This paper further explores the impact of reference sample sizes on model performance, and compares it with popular ML methods. As shown in Figure \ref{fig:preference_chain_vs_ml}, the Preference Chain consistently outperforms the LLM approach, even when only a small amount of reference data is available. From the perspective of KLD, the Preference Chain demonstrates superior performance compared to other methods when the reference data size is less than 100. However, beyond 100 samples, the performance of the MLP begins to surpass that of the Preference Chain.

From the perspective of MAE, the Preference Chain improves with data up to 50, after which it exhibits fluctuation without clear further improvement. Meanwhile, MLP's accuracy gradually surpasses the Preference Chain after 50 samples. These findings suggest that the Preference Chain performs well with limited data, but its advantage decreases as data size increases. A data size of 50 to 100 appears to offer a better balance between performance and cost. It is also important to note that the optimal reference data size may vary depending on the specific task and data characteristics. Therefore, in practical applications, selecting an appropriate reference data size based on the real condition is crucial for achieving the best performance. To further explore the performance of different LLMs and the impact of different factors, more results can be found in the Appendix \ref{appendix:more_results}.

\begin{figure}[!ht]
\centering
    \includegraphics[width=\linewidth]{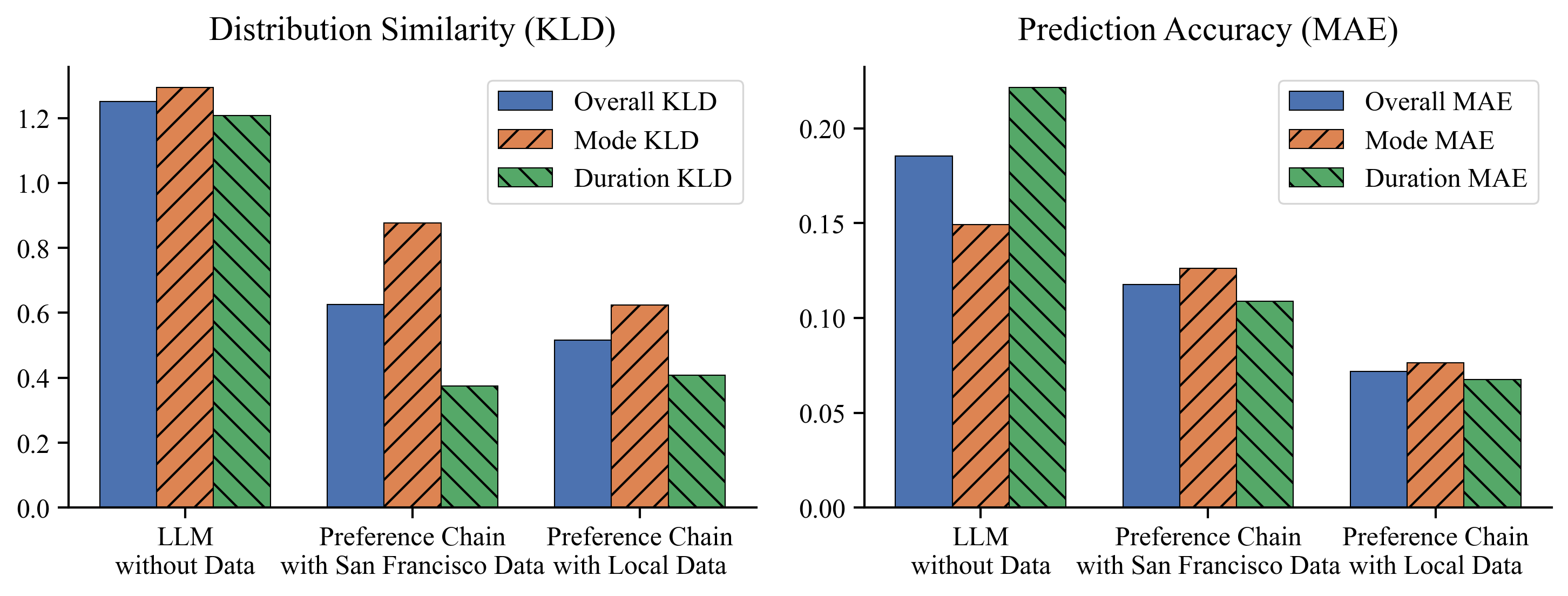}
    \caption{Simulating mode and duration choices in Cambridge using San Francisco's data as reference}
\label{fig:cambridge_simulation}
    \includegraphics[width=\linewidth]{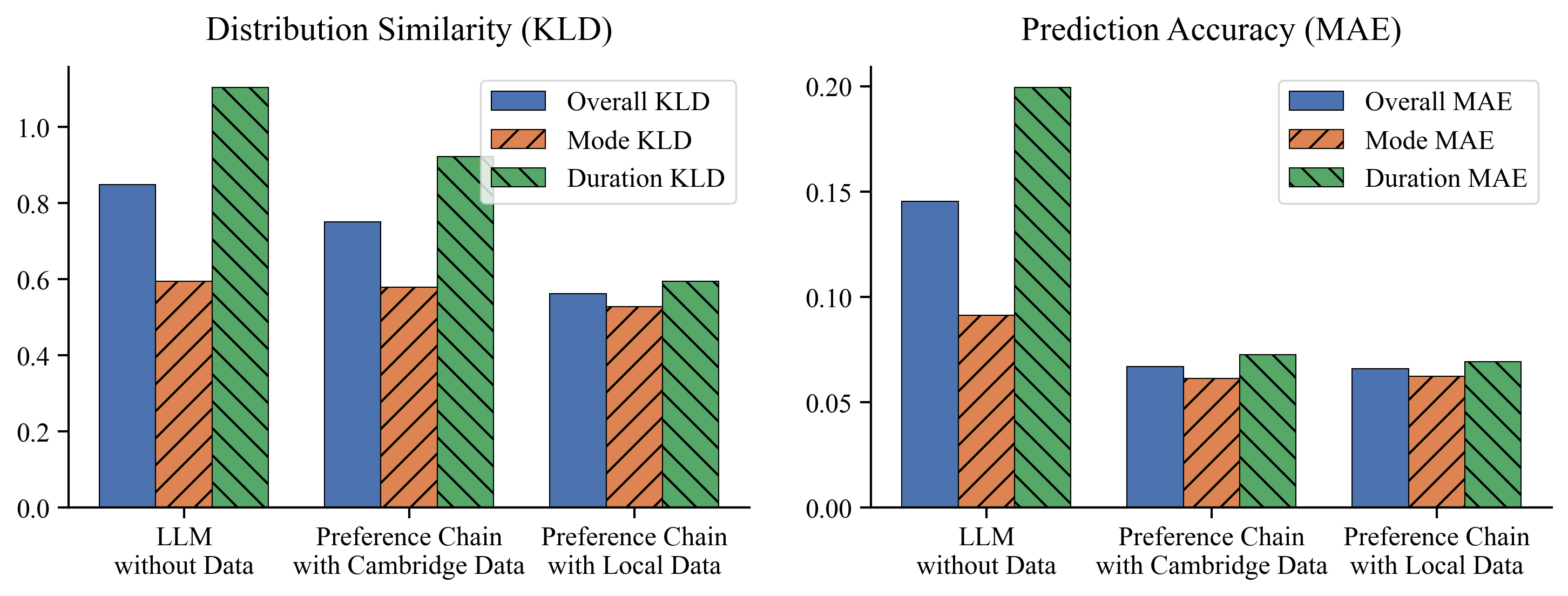}
    \caption{Simulating mode and duration choices in San Francisco using Cambridge's data as reference}
\label{fig:sanfrancisco_simulation}
\end{figure}

\textbf{3. How well will the model perform in other urban environments with different characteristics?}

In practice, urban researchers often face the challenge of having access to open data from certain regions but lacking data for a specific study area. To address this challenge, this paper evaluates the generalizability of the proposed method using reference data from other cities. Figure \ref{fig:cambridge_simulation} shows the experimental results of simulating the transport mode and duration choices in Cambridge, while using reference data from San Francisco. Similarly, Figure \ref{fig:sanfrancisco_simulation} presents the results in San Francisco using Cambridge's data. Both experiments use 1000 samples as validation and 50 samples as reference for the Preference Chain. The experimental results from both regions demonstrate that the use of reference data from other regions can still enhance the performance of LLMs. These findings suggest that, in newly developed areas where data is limited or unavailable, a Preference Chain approach based on external data can still enable more effective behavior simulation.However, it is worth noting that the performance achieved using external data is still inferior to that achieved using local data.

\section{Experiment II: Traffic \& POI Visit Simulation}
\label{sec:exp_2}

\subsection{Mobility Agent}

\begin{figure*}[!ht]
    \centering
    \includegraphics[width=\linewidth]{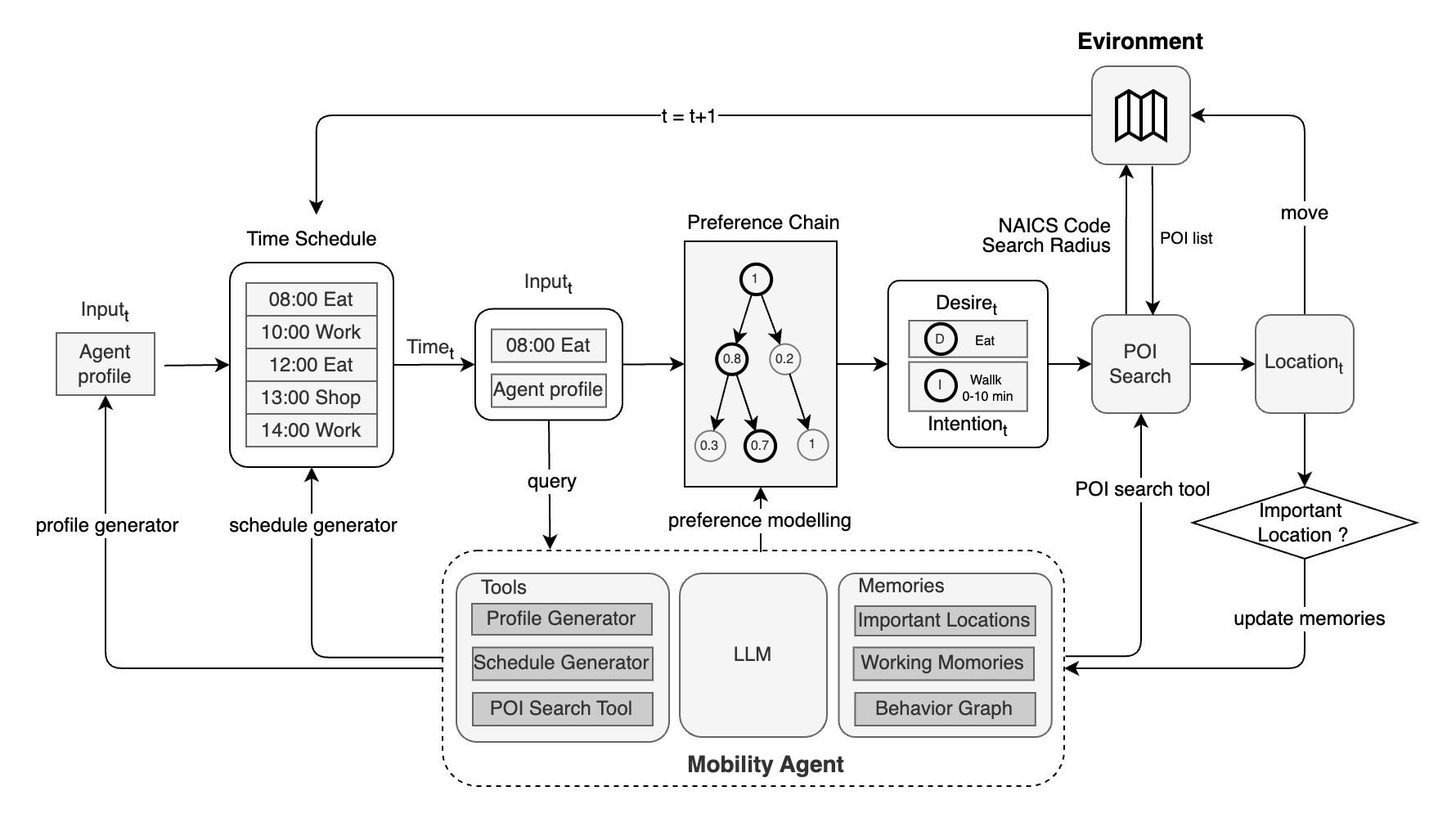}
    \caption{The simulation processes of the Mobility Agent}
    \label{fig:mobility_agent}
\end{figure*}
\begin{figure*}[!ht]
    \centering
    \includegraphics[width=0.8\linewidth]{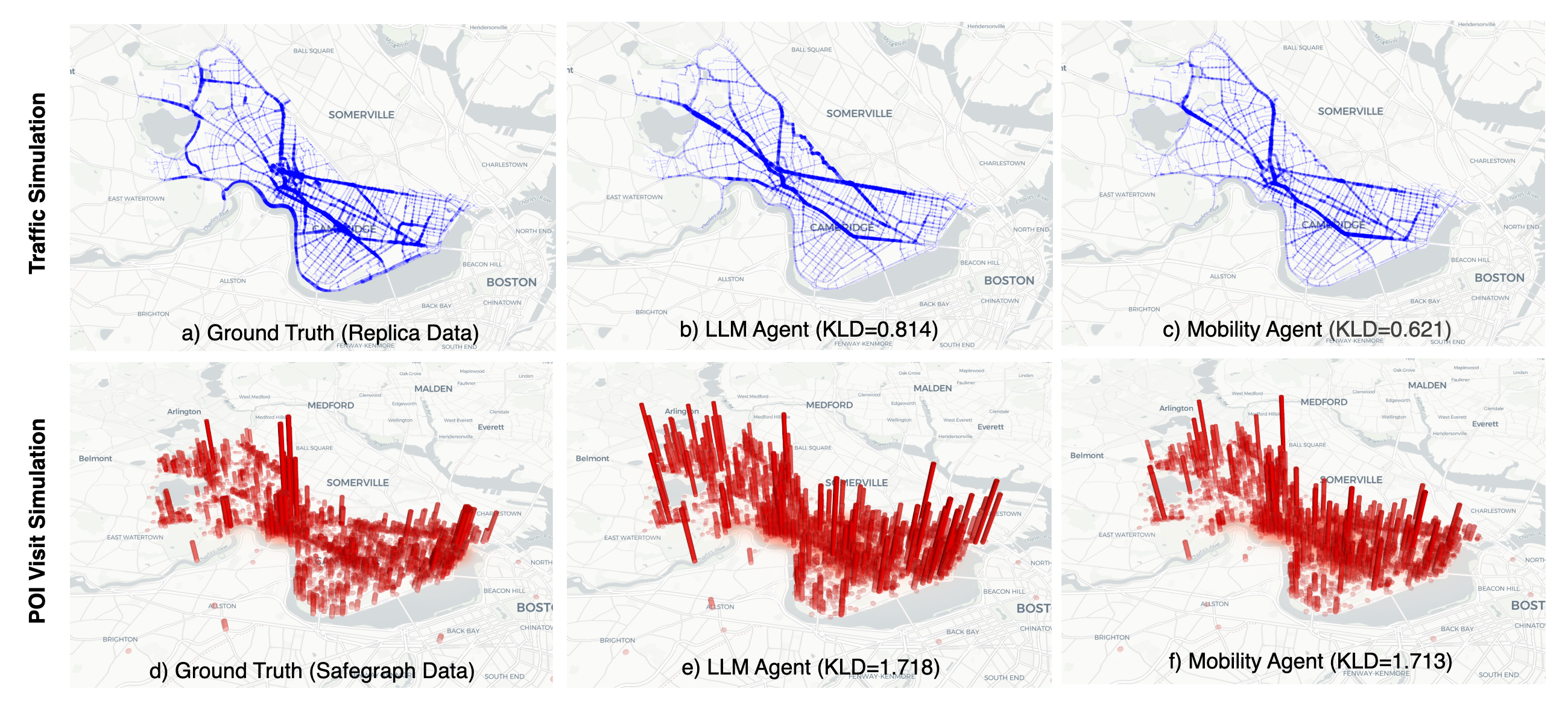}
    \caption{The comparison of using Mobbility Agent and LLM Agent to simulate 1000 people living in the city for 24 hours.}
    \label{fig:abm_simulation}
\end{figure*}

To further explore the application of the proposed method in transportation systems. This paper introduces a traffic simulation agent, named Mobility Agent, with the proposed method. This agent can autonomously plan daily activities, search for nearby POIs, and choose different modes of transportation to travel in the city. An ABM system is built upon the Mobility Agent to simulate 1000 people living in Cambridge, MA for 24 hours. Unlike traditional ABMs, all action logic of the agent is generated by a LLM with Preference Chain, enabling the simulation of more natural and complex urban systems even under data-scarce conditions.

The Mobility Agent have access to three tools: Profile Generator, Schedule Generator and POI Search Tool. Figure \ref{fig:mobility_agent} illustrates the simulation process of the Mobility Agent, which can be divided into the following steps:

\begin{enumerate}
    \item Generating agents with different profiles based on real world demongrapic data using Profile Generator tool.
    \item Generating daily itineraries for different agent using Schedule Generator tool.
    \item Predicting the transportation mode and duration of travel for different agent using the Preference Chain approach.
    \item Using the POI Search Tool to Search for nearby POIs based on travel demand (NAICS Code), mode of transportation, and travel duration.
    \item The LLM finally determines and selects one of the POIs to move to.
    \item Repeating steps 3, 4, and 5 until the simulation of the entire day is completed.
\end{enumerate}

During the simulation, the agent automatically stores important locations, such as home, workplace, and school. The home address is randomly assigned during the agent's initialization, while other locations such as workplaces and schools are autonomously selected and stored by the agent during the simulation process according to its profile. Therefore, the number and types of important addresses vary among different agents, which enhances the fidelity of the simulation to real-world scenarios. In addition, the agent stores important information in its memory, such as dates, weather conditions (if applicable), and previously simulated activities. These pieces of information dynamically influence the agent's decision-making process.

\subsection{Traffic Simulation}

As shown in Figures \ref{fig:abm_simulation} (a–c), this study simulates the 24-hour activities of 1,000 individuals with different profiles in Cambridge,MA using both the LLM Agent and the Mobility Agent based on the Preference Chain (with 50 reference samples). Figure \ref{fig:abm_simulation} (a) presents the total traffic flow for 24 hours in a typical Thursday during Spring, 2024 in the Cambridge area, according to the Replica dataset. It can be observed that the simulation results of the Mobility Agent are more consistent with the real data compared to those of the LLM Agent, as reflected by the KLD metric, which decreases from 0.814 to 0.621 (with lower KLD values indicating better performance). The Mobility Agent successfully identifies Massachusetts Avenue as the primary route, whereas the LLM Agent tends to regard Cambridge Street as the primary route. Additionally, since neither simulation method accounts for external traffic in the Cambridge area, both fail to identify the main traffic corridors along the river.

\subsection{POI Visit Simulation}

Figures \ref{fig:abm_simulation} (d–f) illustrate the normalized number of visits made by the 1,000 agents with different attributes to various POIs throughout 24 hours in a typical workday. Figure 6-14 (d) represents the normalized number of visits to different POIs according to the SafeGraph\cite{safegraph2025} dataset in July 2023. From the results of the POI visit simulation, it appears that the simulation results of both the LLM Agent and the Mobility Agent show similar KLD values. It might due to the lack of direct POI preference information in the reference data. However, the results generated by the Mobility Agent show a more centralized distribution, likely due to the constraints imposed by travel mode choices, which limit the range of POI selections. This characteristic results in a closer alignment with real-world observations.

Overall, this experiment demonstrates that the application of the Preference Chain allows the Mobility Agent to simulate realistic urban mobility by capturing individual preferences. The simulation results show that the Mobility Agent outperforms a standard LLM agent in terms of alignment with real-world traffic behavior. The proposed method is particularly effective in scenarios where data is limited, such as in emerging cities, rural areas or under conditions of sparse data collection. This makes it a valuable tool for research on mobility patterns in data-scarce environments, enabling more accurate and personalized mobility modeling even with minimal input.

\section{Discussion and Conclusion}
\label{sec:discussion}

Simulating human behavior in urban environments presents a promising approach for informed decision-making in urban studies. Traditional ABMs rely on predefined rule sets to simulate human behavior, which limits their capacity to model the uncertainty and variability inherent in human choices. While ML and DL methods offer improvements, these methods either lack flexibility when adapting to changing conditions or demand extensive datasets. State-of-the-art LLMs reveal the potential of generative agents to simulate complex human behavior without relying on additional data. However, limitations remain in modeling behavior in a context-sensitive manner.

To address these limitations, this paper proposes a novel approach that combines Graph RAG with LLMs to achieve more flexible, contextually-aware human behavior modeling. This approach improves the accuracy of LLM approaches, providing reasonable simulations of how different population groups might behave under various conditions. 

\subsection{Key Features and Limitations of the Proposed Method}

The key advantages of the proposed method are outlined below:

\begin{itemize}
    \item \textbf{Effective fuzzy probabilistic prediction}: The proposed method demonstrates good performance in generating probabilistic outcomes that capture the natural uncertainty in human behavior. 

    \item \textbf{Robustness with limited data}: This approach is particularly effective in data-scarce environments, which is valuable for urban studies where large, comprehensive datasets are often unavailable. 

    \item \textbf{Adaptable transfer learning}: The method supports adaptable transfer learning through prompt guidance, enabling simulations across various scenarios. The experiment results demonstrated the effectiveness of proposed method in different urban environment. This flexibility makes it particularly useful for urban simulations with changing or diverse parameters.

    \item \textbf{No training required}: This approach requires no additional training. This feature is particularly advantageous when datasets are continuously updated, allowing for more dynamic and responsive human behavior simulations that can adapt to real-time changes in urban environments.
\end{itemize}

It should also be noted that the proposed method does have limitations:

\begin{itemize}
    \item \textbf{Slow inference}: Since the primary reasoning process relies on an LLM, the inference speed is significantly slower than traditional methods. Although ongoing research aims to optimize LLM inference speed\cite{chopra2024limitsagencyagentbasedmodels}, this remains a major bottleneck.
    
    \item \textbf{Risk of hallucination}: Another inherent risk of LLMs is the potential for generating misleading or false information. Although the proposed method leverages Graph RAG to reduce the hallucination, this risk still exists. The proposed method might not be suitable in critical applications where accuracy is paramount.

    \item \textbf{Discrete model}: The proposed method functions as a discrete model, making it unsuitable for applications requiring continuous predictions. This limitation restricts effectiveness in scenarios where human behavior or environmental factors evolve continuously.
\end{itemize}

\subsection{Potential Applications in Transportation Systems}

The proposed method, which integrates the Preference Chain with LLMs, offers a versatile and context-aware framework for simulating human behavior in transportation systems. This approach is particularly well-suited for scenarios where traditional data-driven models face limitations due to insufficient or incomplete datasets. Below are several potential applications in transportation systems:
\begin{itemize}
    \item \textbf{Urban Mobility Modeling in Data-Scarce Environments}: The method demonstrates strong performance in simulating transportation behavior even with limited data, making it highly applicable in emerging cities, rural areas, or regions with sparse data collection. For instance, transportation planners can use data from well-documented cities to simulate travel patterns in less-studied regions. This capability is crucial for understanding mobility trends in newly developed areas or for evaluating the impact of policy interventions in one region on another.
    \item \textbf{Personalized Travel Behavior Simulation}: The method supports the simulation of diverse population groups based on demographic and socioeconomic characteristics. This allows for the development of personalized mobility models, which can be used to assess the impact of transportation policies on different user groups. Such models are essential for designing equitable and inclusive transportation systems.
    \item \textbf{Dynamic Traffic Simulation and Forecasting}: When integrated with ABMs, the Mobility Agent can simulate daily travel patterns and traffic flows. This approach not only enhances the accuracy of traffic forecasting and congestion analysis, but also enables the modeling of unexpected events such as accidents, road closures, and sudden demand fluctuations. The ability to dynamically adjust to changing conditions makes the method suitable for real-world applications such as smart city initiatives and adaptive traffic control systems.
\end{itemize}

Overall, this paper offers a novel approach for simulating human behavior in transportation systems studies, particularly valuable in informal settlements or rapidly changing neighborhoods where comprehensive datasets are limited. The proposed Preference Chain method demonstrates promising performance in generating natural human behaviors, adapting to diverse scenarios, making it highly suitable for dynamic urban studies. However, challenges such as slow inference speed, the risk of hallucination, and the discrete nature of the model must be addressed in future research. Despite these limitations, this work contributes to the advancement of LLM agents in transportation systems,with potential applications in urban mobility studies, personalized travel simulation, and dynamic traffic forecasting.  It offers a valuable tool for informed decision-making in complex and evolving environments. 

\section*{Code}
\label{sec:code}
The code to reproduce the figures and explore additional settings is available in the following repository: \url{https://github.com/kekehurry/mobility_agent}

\bibliographystyle{unsrt}  
\bibliography{references} 

\onecolumn

\appendix

\section*{Appendix}
\subsection*{Data Schema} \label{appendix:data_schema}

\begin{table}[!h]
\centering
\caption{The data schema of categorized Replica data used in this paper}
\label{tab:data_schema}
\begin{tabular}{p{2cm}p{3cm}p{6cm}}
\hline
\textbf{Type} & \textbf{Name} & \textbf{Categories} \\
\hline
\multirow{7}{*}{Input} & age\_group & Under 18, 18-24, 25-34, 35-44, 45-54, 55-64, 65+ \\
& income\_group & Under \$10k, \$10k-\$50k, \$50k-\$100k, \$100k-\$150k, \$150k-\$200k, \$200k-\$300k, \$300k+ \\
& employment\_status & under\_16, not\_in\_labor\_force, unemployed, employed \\
& household\_size & 1, 2, 3, 4, 5, 6, 7, 8 \\
& available\_vehicles & zero, one, two, three\_plus, unknown\_num\_vehicles \\
& education & no\_school, k\_12, high\_school, bachelors\_degree, advanced\_degree, some\_college \\
& trip\_purpose & eat, work, home, school, shop, maintenance, social, recreation, other\_activity\_type \\
& start\_time & 0, 1, 2, ..., 23 \\
\hline
\multirow{2}{*}{Output} & primary\_mode & walking, biking, auto\_passenger, public\_transit, private\_auto, on\_demand\_auto, other\_travel\_mode \\
& duration\_minutes & 0-10, 10-20, 20-30, 30-40, 40-50, 50-60 \\
\hline
\end{tabular}
\end{table}

\begin{figure}[!h]
    \centering
    \includegraphics[width=0.8\linewidth]{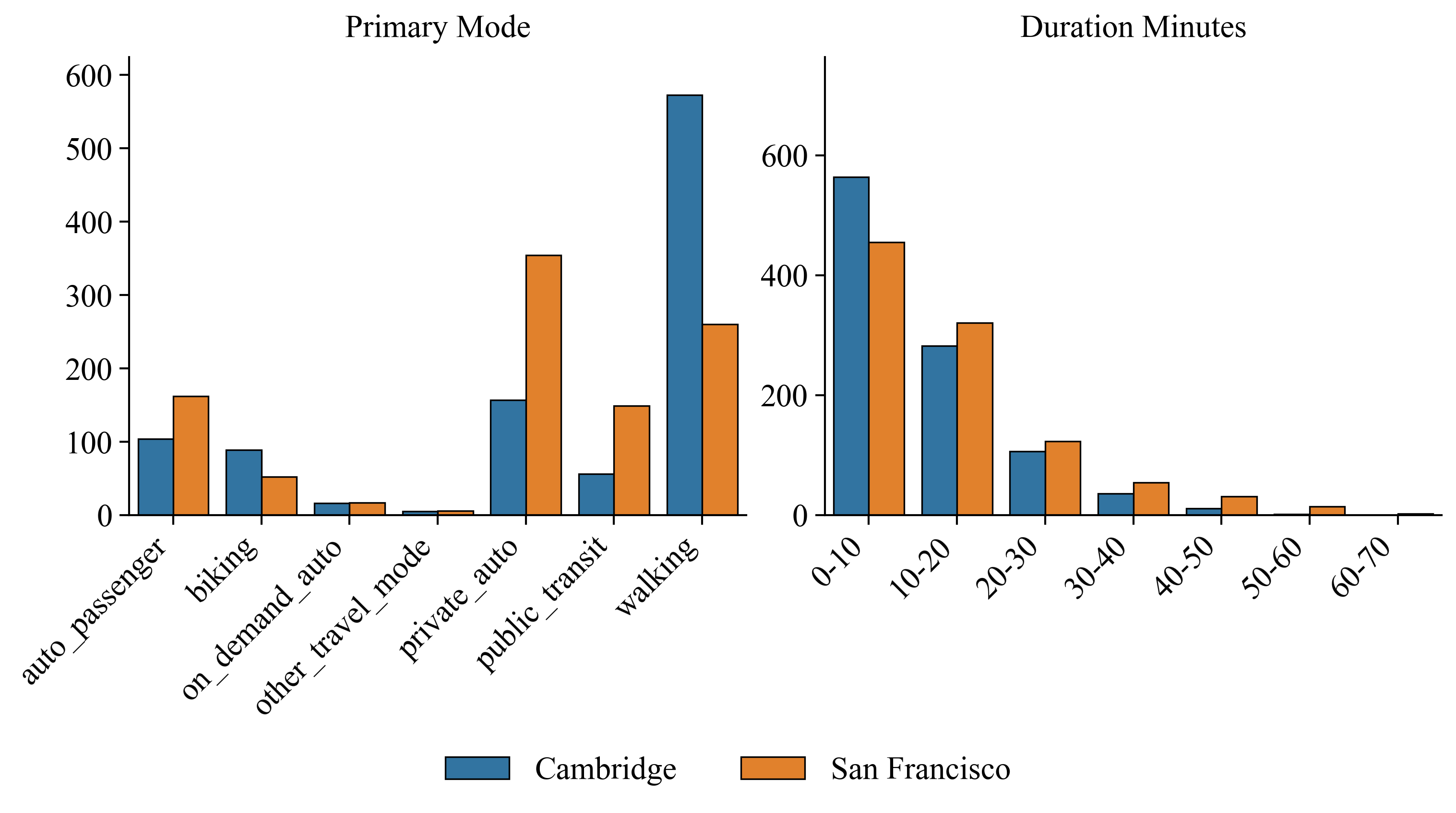}
    \caption{The distribution of output features in the dataset}
    \label{fig:data_output_distributions}
\end{figure}

\begin{figure}[!h]
    \centering
    \includegraphics[width=0.8\linewidth]{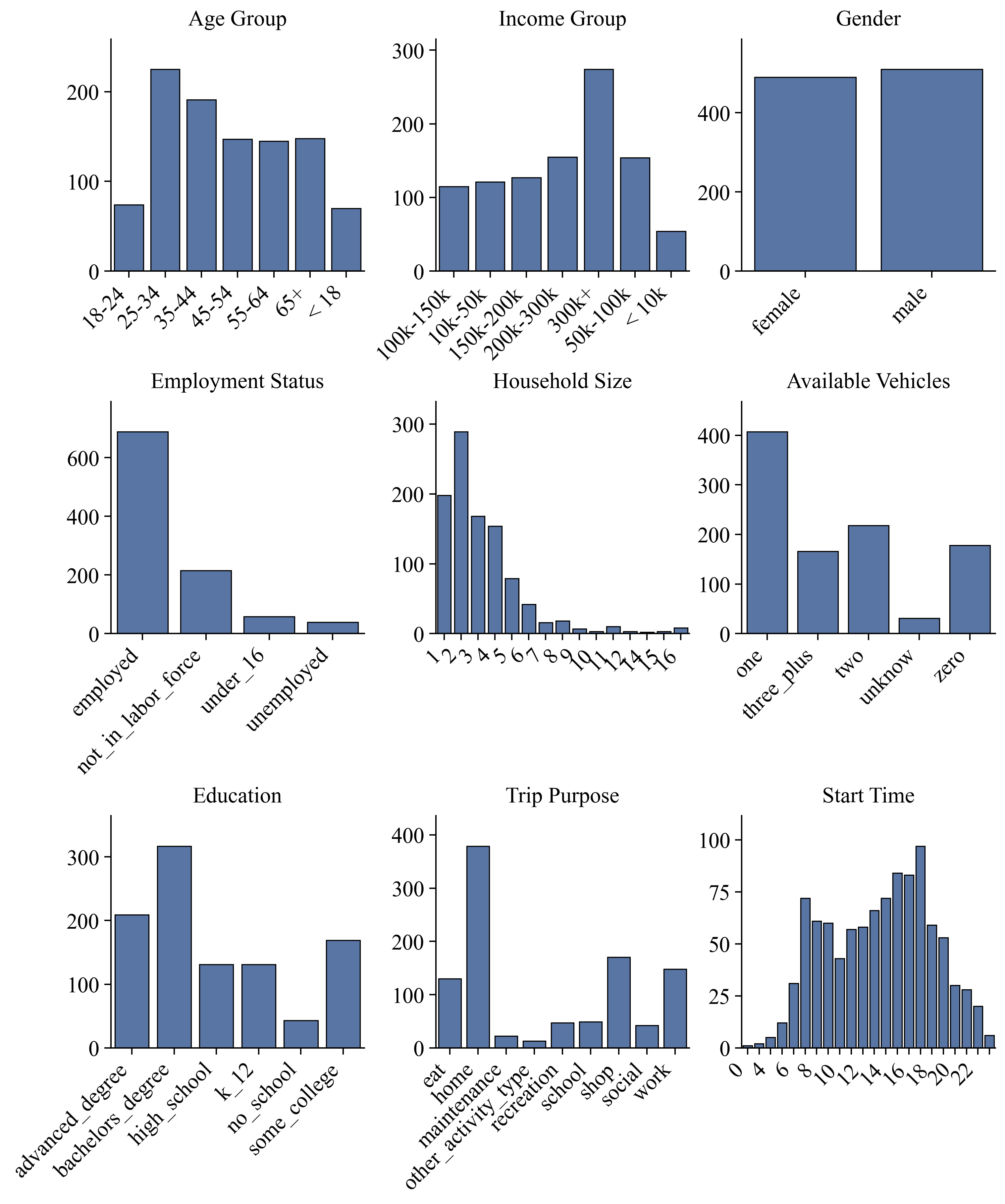}
    \caption{The distribution of input features in the San Francisco dataset}
    \label{fig:data_sanfransico}
\end{figure}

Table \ref{tab:data_schema} presents the data schema used in this paper, derived from the Replica dataset. The input variables include demographic and trip-related features such as age group, income group, employment status, household size, number of available vehicles, education level, trip purpose and trip start time. The output variables consist of the primary mode of travel and trip duration in minutes, both of which are categorized for modeling purposes. Figure \ref{fig:data_output_distributions} plots the output distributions. And Figure \ref{fig:data_sanfransico} demonstrates the distributions of input features in San Francisco.  

\subsection*{Compared Models} \label{appendix:compared_models}

\begin{table}[!h]
\centering
\caption{Configuration of compared machine learning models}
\label{tab:compared_models}
\begin{tabular}{p{2cm}p{3cm}p{6cm}}
\hline
\textbf{Model} & \textbf{Description} & \textbf{Parameters} \\
\hline
RF & Python sklearn implementation of Random Forest algorithm & 
\begin{itemize}
\item n\_estimators: [10, 50, 100]
\item max\_depth: [None, 10, 20]
\item min\_samples\_split: [2, 5, 10]
\item min\_samples\_leaf: [1, 2, 4]
\end{itemize} \\
\hline
MLP & Python sklearn implementation of Multi-layer Perceptron neural network & 
\begin{itemize}
\item hidden\_layer\_sizes: [(64,64), (128,128), (256,256)]
\item activation: ['tanh', 'relu', 'logistic']
\item alpha: [1e-4, 1e-3, 1e-2]
\end{itemize} \\
\hline
XGB & Official implementation of XGBoost algorithm & 
\begin{itemize}
\item n\_estimators: [10, 50, 100]
\item max\_depth: [None, 10, 20]
\item learning\_rate: [1e-5, 1e-4, 1e-3]
\item subsample: [0.7, 0.8, 0.9]
\item gamma: [0, 0.1, 0.2]
\end{itemize} \\
\hline
Qwen3:8b & An open-sources LLM created by Alibaba Cloud & 
\begin{itemize}
\item Model: Qwen3-8B
\item Top-k: 20
\item Temperature: 0.6
\item Top-p: 0.95
\item Repeat Penalty 1
\end{itemize} \\
\hline
Preference Chain & The proposed method & 
\begin{itemize}
\item Model: Qwen3-8B
\item Top-k: 20
\item Temperature: 0.6
\item Top-p: 0.95
\item Repeat Penalty 1
\end{itemize} \\
\hline
\end{tabular}
\end{table}

The ML models compared with proposed method in section \ref{sec:exp_1} utilize grid search to determine optimized parameters. Table \ref{tab:compared_models} presents the configuration of each model. The detailed training process can be found in the repository mentioned in section\label{sec:code}. The LLM and the Preference Chain methods use Qwen3:8b without thinking mode, and the same generation parameters are applied for consistency. 

\subsection*{More Results} \label{appendix:more_results}

\textbf{1. Prompts and example responses}

\begin{figure}[!h]
    \centering
    \includegraphics[width=0.5\linewidth]{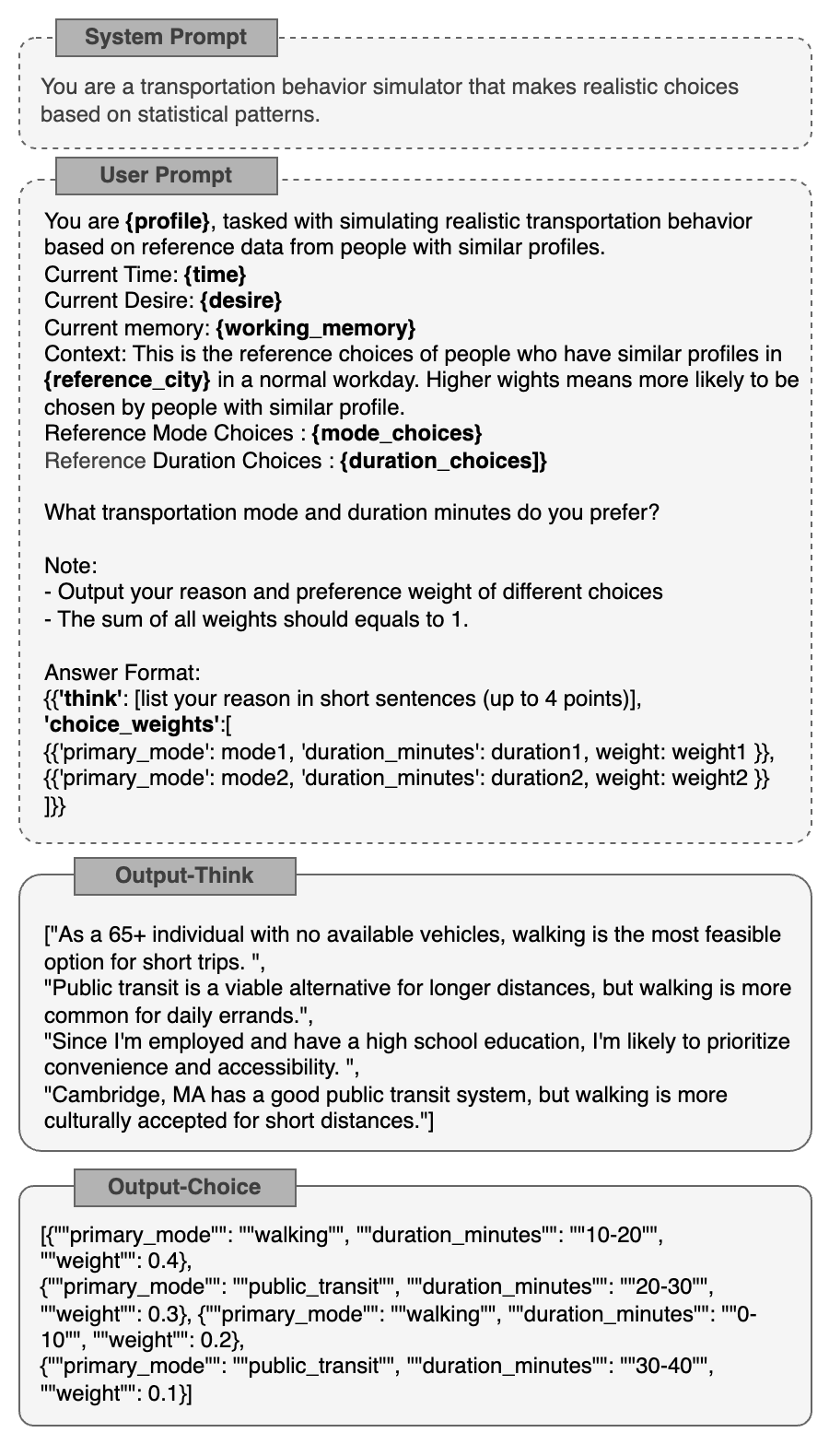}
    \caption{The key prompts and example responses of proposed method.}
    \label{fig:example_reponse}
\end{figure}

Figure \ref{fig:example_reponse} demonstrates the key prompts used in the paper to guide the LLMs generate transportation mode choices. The reference mode choices and duration choices and their weights are calculated by the Preference Chain method mentioned in the section \ref{sec:methodology}.

\textbf{2. Comparison between different LLMs}

\begin{figure}[H]
    \centering
    \includegraphics[width=0.8\linewidth]{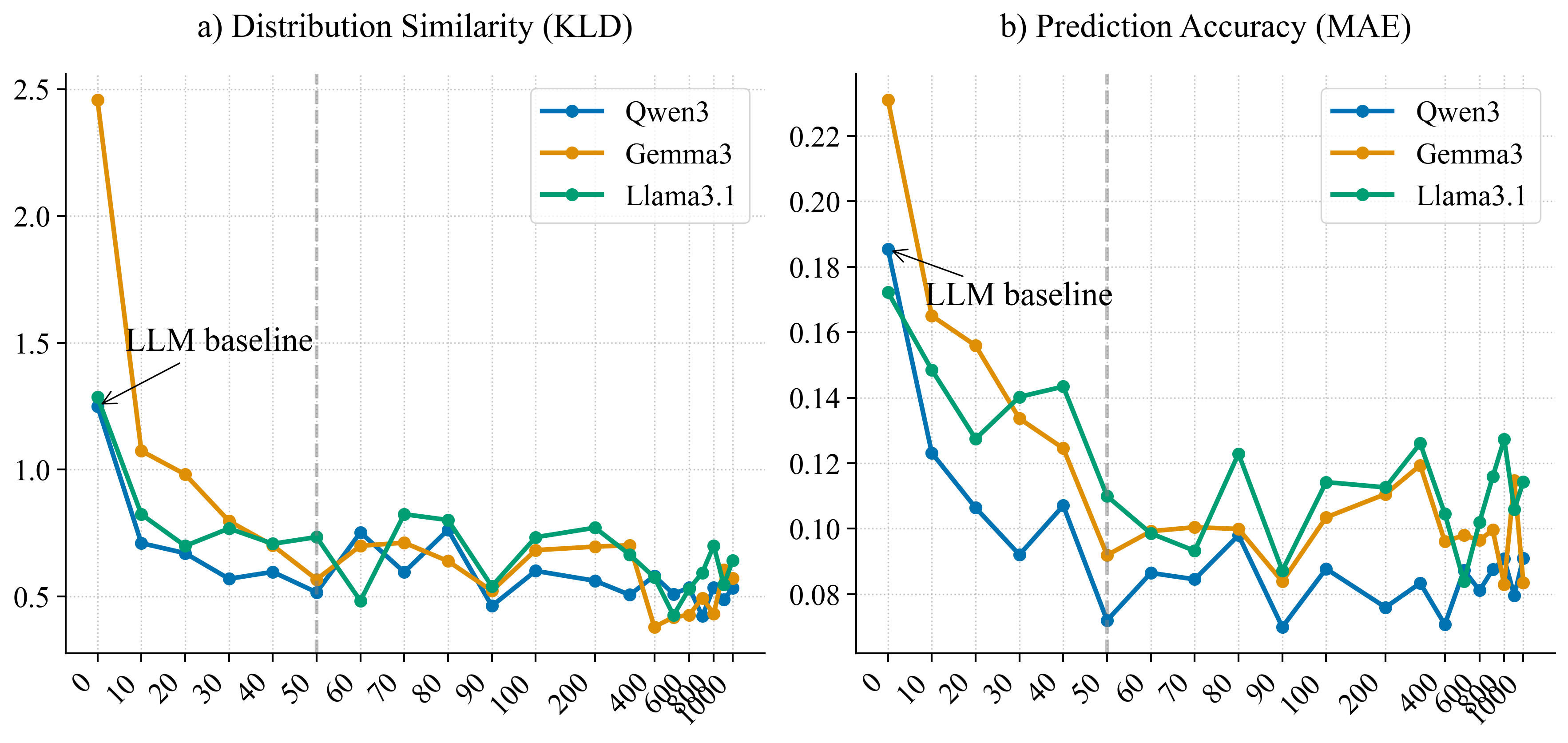}
    \caption{Comparision of the model performance based on different LLMs.}
    \label{fig:different_llms}
\end{figure}

To investigate the performance of the Preference Chain method across different LLMs, this paper conducted comparative experiments using open-source models from different companies: Qwen3:8b, Gemma3:12b, and Llama3.1:8b, as shown in the Figure \ref{fig:different_llms}. From the baseline performance without reference data, the simulation results of Llama3.1:8b show better alignment with real human behavior. When applying the Preference Chain method, Qwen3:8b has the best performance. Meanwhile, the performance of different LLMs exhibits similar trends: when the amount of reference data is less than 50, the model performance improves significantly with the increase in reference data. However, when exceeds 50, the performance gain from further increasing the data size becomes less significant. It indicate that the proposed method achieves higher cost-effectiveness in improving the performance of LLMs in simulating human behavior when using a relatively small amount of reference data.

\textbf{3. Semantic analysis of LLM thinking output}

\begin{figure}[H]
    \centering
    \includegraphics[width=0.8\linewidth]{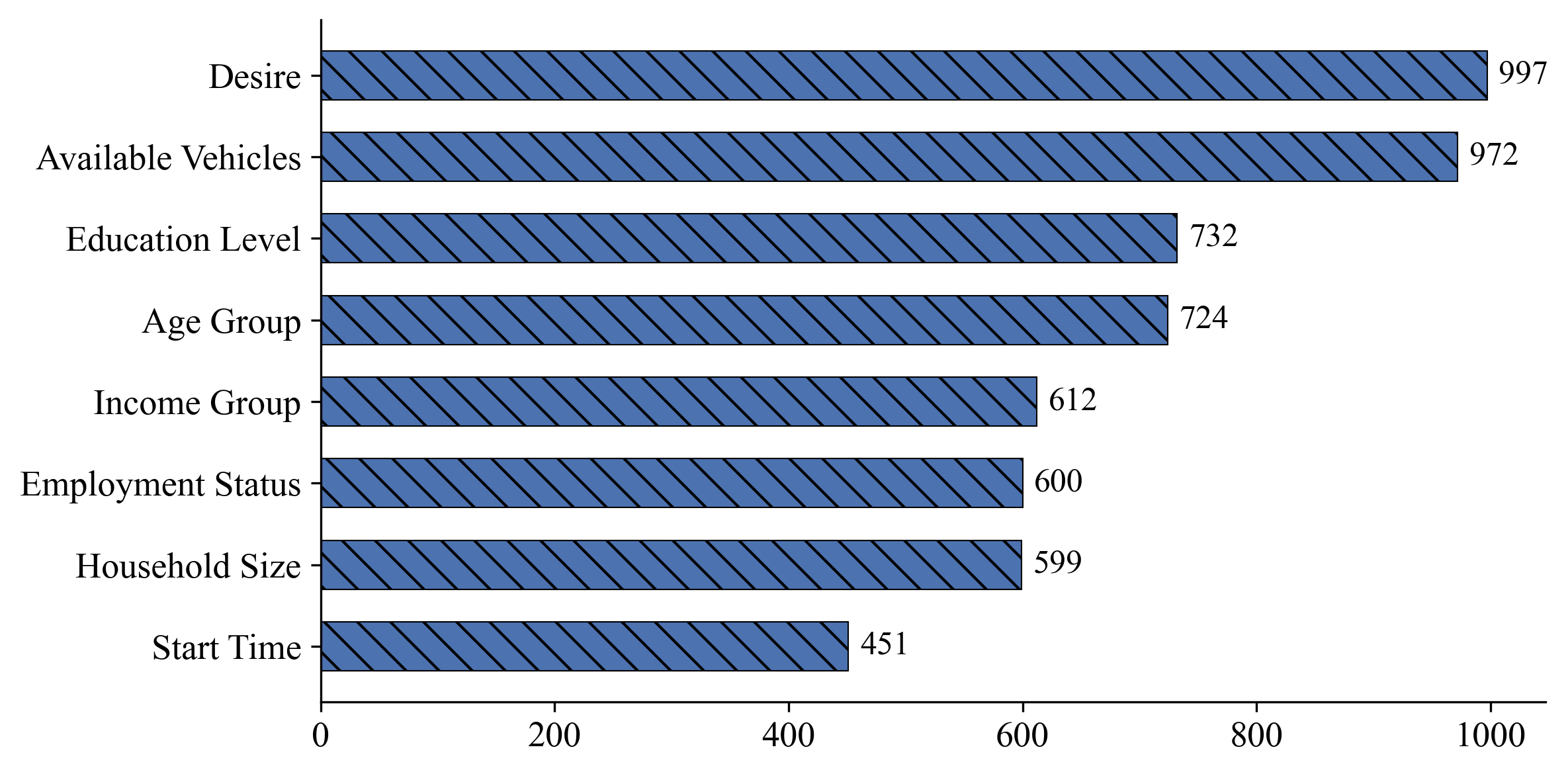}
    \caption{The frequency of different factors present in the LLM thinking outputs}
    \label{fig:factor_importance}
\end{figure}

\begin{figure}[H]
    \centering
    \includegraphics[width=0.8\linewidth]{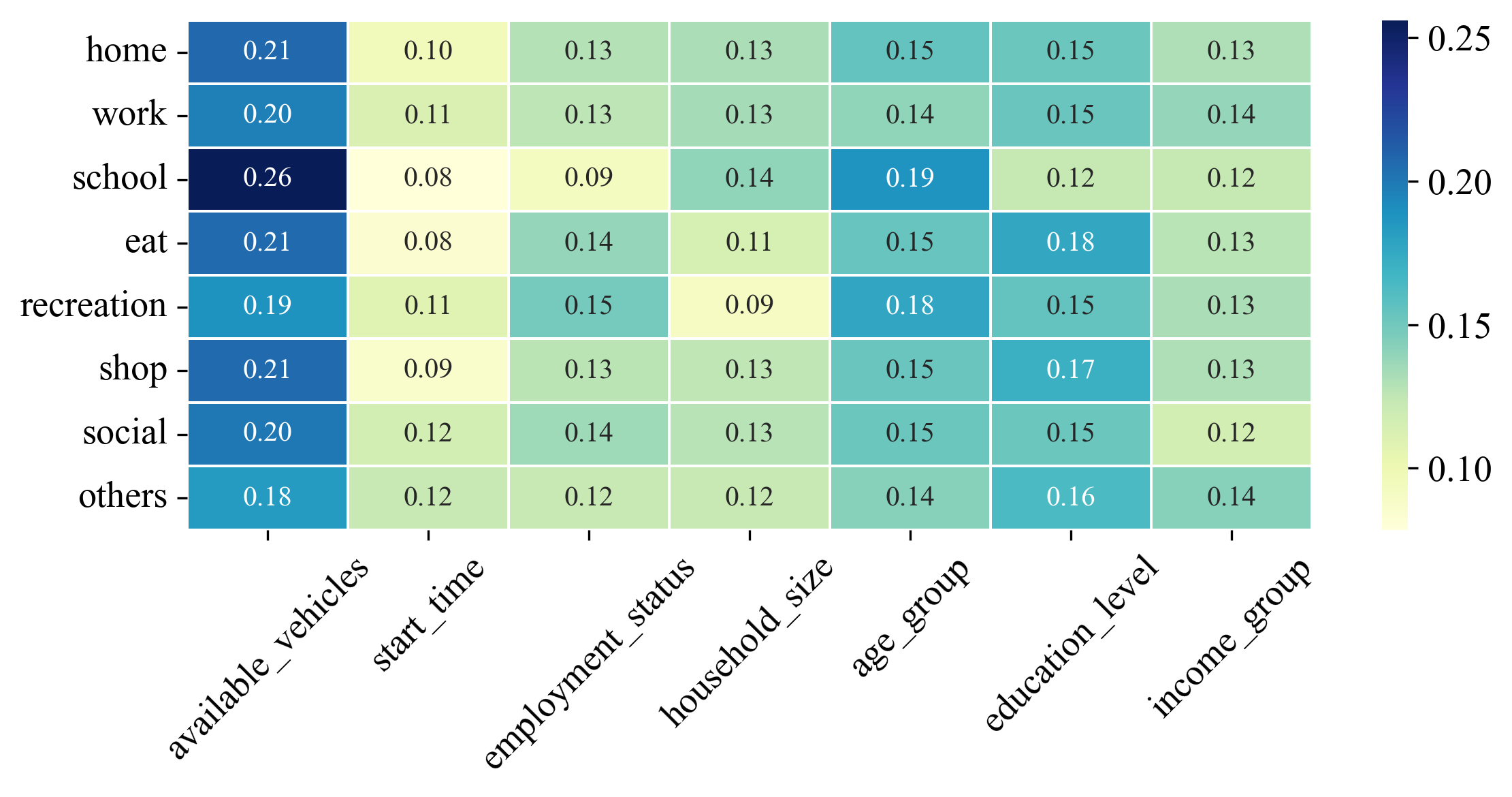}
    \caption{Normalized frequency of different factors across diverse desires}
    \label{fig:factor_analysis_across_desires}
\end{figure}

As shown in Figure \ref{fig:factor_importance},\ref{fig:factor_analysis_across_desires}, the LLM thinking output is categorized into different factors using Qwen3:8b. By quantifying the frequency of each factor considered by the LLM, this paper evaluate their relative importance in the decision-making process of the model. The results indicate that the model tends to prioritize factors related to current desire and immediate availability when formulating its responses, while demographic and socioeconomic factors are also considered.

\end{document}